\documentclass[10pt, a4paper]{article}

\usepackage[final]{lrec2026} % this is the new style
\usepackage{graphicx}
\usepackage{booktabs}
\usepackage{pifont}
\usepackage{booktabs}
\usepackage{amsmath}
\usepackage{enumitem}
\usepackage{xcolor}
\usepackage{float}
\usepackage{multirow}
\usepackage{array}
\usepackage{subcaption}
\usepackage{makecell}

\usepackage{todonotes}

\title{Using Multimodal and Language-Agnostic Sentence Embeddings for Abstractive Summarization
}

\name{Chaimae Chellaf\textsuperscript{1,2}, Salima Mdhaffar\textsuperscript{1}, Yannick Estève\textsuperscript{1}, Stéphane Huet\textsuperscript{1}} 

\address{\textsuperscript{1}LIA - Avignon Université, France\\
         \textsuperscript{2}Lundi Matin, France \\
         {chaimae.chellaf-el-hammoud@alumni.univ-avignon.fr}\\ }
\abstract{
Abstractive summarization aims to generate concise summaries by creating new sentences, allowing for flexible rephrasing.
However, this approach can be vulnerable to inaccuracies, particularly `hallucinations' where the model introduces non-existent information.
In this paper, we leverage the use of multimodal and multilingual sentence embeddings derived from pretrained models such as LaBSE, SONAR, and BGE-M3, and feed them into a modified BART-based French model. A Named Entity Injection mechanism that appends tokenized named entities to the decoder input is introduced, in order to improve the factual consistency of the generated summary.
Our novel framework, SBARThez, is applicable to both text and speech inputs and supports cross-lingual summarization; it shows competitive performance relative to token-level baselines, especially for low-resource languages, while generating more concise and abstract summaries.
 \\ \newline  }

\begin{document}

\maketitleabstract

\section{Introduction}
Automatic text summarization plays a critical role in distilling vast amounts of information into concise and informative outputs~\cite{Zhang2025}.
%Extractive summarization, which preserves exact wording by selecting and reordering existing text, has long been the primary focus of research.
With the advent of deep learning, neural approaches have become the foundation of modern summarization systems, leveraging large-scale training data and Transformer-based architectures \cite{vaswani2017attention} to model contextual relationships across sequences of text. These models, such as BART \cite{lewis2020bart}, T5 \cite{raffel2020exploring}, and more recently, large language models (LLMs) like GPT \cite{radford2018improving, brown2020language} generate summaries by operating at the token level, predicting the next token conditioned on both the input tokens and the previously generated ones. 
While these models perform well in abstractive summarization, they still exhibit notable limitations that impact the overall quality of the summaries \cite{shakil2024abstractive}.
Extractive summarization focuses on selecting the most important sentences or phrases directly from the source text to form a summary without generating new material.
In contrast, abstractive summarization is a more sophisticated technique that produces new sentences to convey the main ideas of the source text; instead of copying, this technique paraphrases and synthesizes the content to present it in a coherent and natural way~\cite{Zhang2025}.
In addition, two important extensions of text summarization are cross-lingual summarization \cite{wang2022survey} and cross-modal summarization \cite{sharma2022end, retkowski2025speech}. 
Cross-Lingual Summarization aims to generate summaries in a language different from the source text, while cross-modal summarization involves different input-output modalities, such as speech-to-text or text-to-speech summarization. 
These approaches expand access to information across languages and modalities.
However, both tasks face significant challenges, such as the scarcity of parallel summarization datasets, and error propagation in pipeline systems \cite{perez2021models, wang2022survey}. For example, speech-to-text summarization is often compromised by transcription errors, while cross-lingual cascade models can be marred by translation errors. 

To address these challenges, \citet{barrault2024large} presented Large Concept Models (LCMs) which operate independently of any instantiation in a particular language or modality.
LCMs propose a shift from the traditional token-level representation of input texts to sentence-level representation. 
Instead of processing input text as discrete tokens, they operate within a continuous sentence embedding space wherein each sentence is encoded as a dense vector capturing its underlying semantic content.
This representation framework facilitates more abstract reasoning over higher-level semantic units, thereby improving conceptual generalization. 

In line with this work, we introduce a novel abstractive summarization model that, unlike LCMs, is designed to work under significantly reduced computational requirements, tailored specifically for the summarization task. 
While LCM targets general-purpose language modeling, our approach focuses on learning abstract summaries using compact architectures trained on task-specific data. Our contributions are:
\begin{itemize}
\item We propose a novel modified BART-based architecture, SBARThez, which performs summarization from sentence embeddings derived from pretrained models such as LaBSE, SONAR, and BGE-M3. In contrast to conventional token-level encoding methods, our architecture breaks new ground by effortlessly processing text data from various languages as well as audio inputs. 
%We will release the training code and model checkpoints in the camera-ready version.
We make the training code of our models freely available at 
\url{https://github.com/cchellaf/SBARThez}
%\url{https://anonymous.4open.science/r/SBARThez-8CAC}
\item We introduce a strategy that incorporates the explicit injection of named entities into the decoder step of the summarization process, aiming to mitigate the risk of hallucinated entities.
\item We conduct extensive evaluations of our architecture on a diverse set of benchmarks spanning multiple language pairs (X to French) and two input modalities (text and speech). Our results demonstrate that despite its smaller scale, it achieves competitive performance compared with several baselines, with notable gains on low-resource languages and superior scores on measures of abstractness.
\end{itemize}

\vspace{-0.22cm}
\section{Related Work}

\subsection{Text Sentence Embedding Extraction}
Recent advancements in multilingual sentence embeddings have significantly improved the performance of cross-lingual and semantic similarity tasks. One pioneering model, Sentence-BERT (SBERT) \cite{reimers2019sentence}, introduced a Siamese network architecture over BERT, enabling efficient computation of semantic similarities between sentence pairs. Initially trained on English, SBERT was later adapted for use in other languages. Another embedding model, LaBSE, introduced by \cite{feng2022language}, uses a novel combination of pretraining and dual-encoder fine-tuning to enhance translation ranking performance. 
It offers robust embeddings across 109 languages and has been widely adopted for tasks requiring language-agnostic representations, such as bitext retrieval and downstream classification. 
More recently, a multilingual and multimodal fixed-size sentence embedding space called SONAR \cite{duquenne2023sonar} uses an encoder-decoder approach to build its sentence embedding space, with training on large-scale text data. 
The SONAR text encoder covers 200 languages, improving both alignment and semantic coherence across these languages. 
SONAR demonstrated strong performance, particularly in zero-shot cross-lingual retrieval. 
Jina Embeddings~\cite{sturua2024jina} further extended multilingual capabilities by offering embedding models optimized for query-document retrieval, clustering, classification, and text matching, achieving competitive performance across multiple benchmarks.
Finally, another notable embedding model, BGE-M3, has emerged as a versatile solution~\cite{chen2024m3}, supporting over 100 languages and excelling across three key dimensions: multilinguality, 
multifunctionality, and multigranularity.
%It can handle inputs of varying lengths, ranging from short sentences to lengthy documents of up to 8,192 tokens.

%A primary use case for these embedding models is neural retrieval, where they enable the identification of semantically relevant answers by comparing text embeddings based on their similarity. These models are also fundamental to Retrieval-Augmented Generation (RAG) \cite{gao2023retrieval}, enhancing language models by retrieving and incorporating relevant information to generate more accurate and contextually rich responses. 

\subsection{Speech Utterance Embedding Extraction}
Compared to earlier approaches, which focused on acoustic frame-level representations (10–20 ms), speech utterance embedding models take a more global approach, learning multimodal and multilingual representations at the speech level (5–10 seconds). 
These models aim to align the resulting embedding space semantically across languages. 
For instance, SAMU-XLSR \cite{khurana2022samu} successfully adapted the XLS-R speech encoder \cite{babu2022xls} to generate utterance-level speech embeddings that rival those of language-agnostic sentence embedding models like LaBSE~\cite{feng2022language}. 
This is achieved through a teacher-student framework, where LaBSE is frozen as the teacher, and the speech encoder is trained to generate embeddings that closely match the textual embeddings using cosine similarity.
A similar approach is taken in SONAR \cite{duquenne2023sonar}, where a pretrained sentence embedding space is used as the teacher. This space is constructed using an 
encoder-decoder model initialized from the NLLB 1B model~\cite{costa2022no}.
The SONAR speech encoders are then trained to align with this shared space, starting from a pretrained w2v-BERT 2.0 model \cite{barrault2023seamless}.
A more recent model, SENSE \cite{mdhaffar2025sensemodelsopensource}, also leverages w2v-BERT 2.0 as the speech encoder and BGE-M3 \cite{chen2024m3} as the teacher model.
These models provide rich, multilingual, and semantically aligned speech representations suitable for downstream tasks like speech summarization.

\vspace{-0.25cm}
\subsection{Abstractive Summarization}

%Abstractive summarization mimics how humans summarize content. It can condense information effectively and is better suited for summarizing texts with varying opinions, perspectives, or complex structures, as it can integrate and rephrase information from multiple parts of the source.
The evolution of neural architectures, particularly sequence-to-sequence (seq2seq) models~\cite{shi2021neural} and transformer-based approaches, has revolutionized abstractive text summarization. 
The availability of large-scale supervised summarization datasets has enabled deep learning techniques to thrive~\cite{grusky2018newsroom, scialom2020mlsum}.
%Earlier methods were based on recurrent networks \cite{nallapati2016abstractive}.
Notable milestones include the introduction of self-attention mechanisms in models like BERT~\cite{devlin2019bert}, BART \cite{lewis2020bart}, T5 \cite{raffel2020exploring}, and PEGASUS \cite{zhang2020pegasus}, which have significantly improved text summarization 
capabilities.
%with their self-attention mechanisms, marked an important step in text summarization techniques, as self-attention allows the models to capture contextual information across the entire sequence.
%More recently, reinforcement learning techniques~\cite{alomari2022deep} have been integrated with these models, allowing them to optimize summary quality based on human-like preferences.
Pretrained large language models, including GPT \cite{radford2018improving}, LLaMA \cite{touvron2023llama}, and Gemini models \cite{team2023gemini}, particularly their instruction-tuned variants, have demonstrated better generalization in many natural language processing tasks, including text summarization. They excel at distilling key information into fluent summaries, as shown on recent benchmarks~\cite{zhang2024benchmarking}.

\vspace{-0.2cm}
\section{Methodology}
%In this section, we detail the methodology adopted in this study. 
%We first describe the architecture of the proposed model in subsection \ref{sub:archi}.
%Next, we outline the pretrained models selected as the foundation of our approach in subsection \ref{sub:pretrained}.
%The evaluation metrics used to assess model performance are presented in subsection \ref{sub:metrics}.
%We then introduce the summarization tasks considered in this work, and associated datasets in subsection \ref{sub:tasks}. 
%Finally, subsection \ref{sub:training} provides the training details of our approach. 

We first describe our model’s architecture in subsection \ref{sub:archi} and the pretrained models on which it builds in subsection \ref{sub:pretrained}.
The evaluation metrics used to assess model performance is introduced in subsection \ref{sub:metrics}.
We then present the summarization tasks and datasets considered in this work in subsection \ref{sub:tasks}. 
Finally, subsection \ref{sub:training} outlines the training details of our approach.

\subsection{Model Architecture}
\label{sub:archi}
\begin{figure}[ht!]
  \centering
  \includegraphics[width=0.8\columnwidth, keepaspectratio]{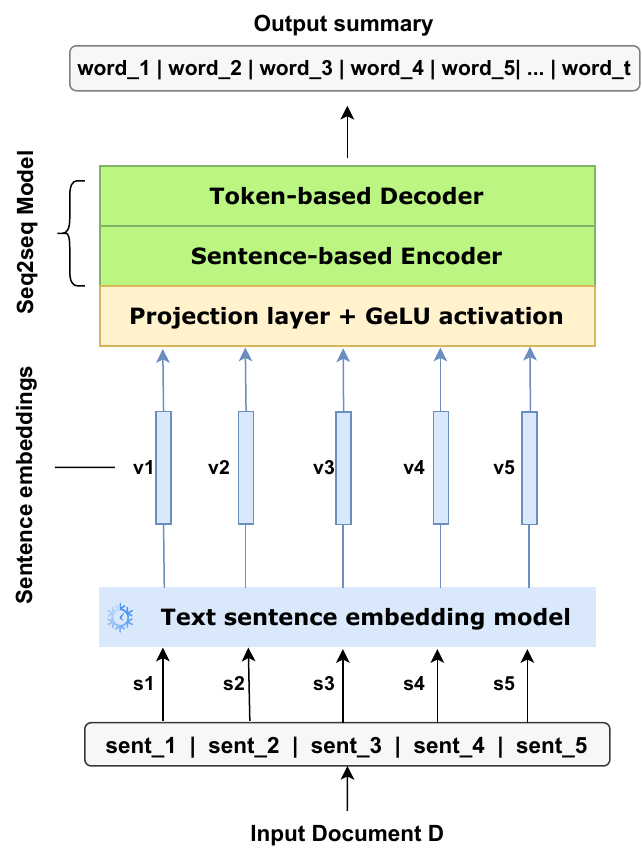}
  \caption{The sentence-based summarization model architecture}
  \label{fig:model_archi_wo_ne}
\end{figure}

We propose an abstractive summarization model that generates summaries based on sentence-level embeddings, as illustrated in Figure~\ref{fig:model_archi_wo_ne}.
The input document \( D \) is first divided into sentences \( D = [s_1, s_2, s_3, \ldots, s_n] \). 
Each sentence \( s_i \) is encoded into a vector \( v_i \) using a text sentence embedding model. 
The resulting sequence of vectors \( V = [v_1, v_2, v_3, \ldots, v_n] \) is then fed into a modified pretrained seq2seq model, which is trained to produce the final summary. 
To guarantee compatibility with the input requirements of the seq2seq model, we incorporate an additional linear projection layer that dynamically adjusts the 
dimensionality of sentence embeddings.
This layer is only applied when the existing embeddings do not conform to the expected input shape of the model.

In this study, we employ a pretrained token-based encoder-decoder seq2seq model and modify its architecture to accept sentence embedding vectors directly.
To achieve this, we remove the encoder’s embed-tokens layer, which generates token-level embeddings.
By doing so, our approach can leverage sentence-level representations, obviating the need for tokenization within the encoder.
As a result, the modified encoder operates 
on entire sentences, while retaining the unchanged decoder architecture and token-based processing.

The training process of our summarization model consists of two stages. 
In the first stage, the model is trained on a large-scale textual summarization dataset to adjust its weights to effectively process sentence embeddings. 
%In the second stage, training continues on a summary dataset specifically designed for evaluation purposes.
In the second stage, the model is fine-tuned on a dataset aligned with the intended summarization task. 
Throughout both stages, the text sentence embedding model remains frozen, and only the projection layer and the modified pretrained seq2seq model are trained. 

%\begin{figure}[t]
%  \includegraphics[width=\textwidth, height=0.3\textheight, keepaspectratio]{latex/figures/SBARThez-sans-barthez2.drawio.png}
%  \caption{Our model's architecture}
%  \label{fig:model_archi}
%\end{figure}

%\begin{figure}[h!]
%  \includegraphics[width=\columnwidth, keepaspectratio]
%  {latex/figures/nouv_schema_sbarthez.drawio2.pdf}
  
%  \caption{Our model's architecture}
%  \label{fig:model_archi2}
%\end{figure} 

%\begin{figure*}[ht!]
%  \centering
%  \includegraphics[width=\textwidth, keepaspectratio]{latex/figures/nouv_schema_sbarthez.drawio2.pdf}
%  \caption{Our model's architecture: The green blocks represent the modified encoder-decoder of the pretrained seq2seq model. The blue blocks remain frozen during training. The orange dashed-line blocks represent an optional module that may or may not be used depending on the setup.}
%  \label{fig:model_archi}
%\end{figure*} 

%\begin{figure}[h!]
%  \includegraphics[width=\columnwidth, keepaspectratio]
%  {latex/figures/SBARThez_NE_injection.drawio.pdf}
%  
%  \caption{The architecture of our modified seq2seq model with NE injection mechanism}
%  \label{fig:model_ner_archi}
%\end{figure} 

%\subsection{Experimental Framework}
%\yk{Ajouter une intro pour les sous-sections suivantes}
\subsection{Pretrained Models} 
\label{sub:pretrained}
\textbf{Text sentence embedding models:} Three embedding models are compared in this study: BGE-M3 \cite{chen2024m3} producing a 1024-dimensional embedding based on the CLS token representation; LaBSE \cite{feng2022language}, which produces a 768-dimensional vector with the same CLS token approach; and SONAR~\cite{duquenne2023sonar}, producing a 1024-dimensional embedding through mean-pooling of its token representations. \\
\textbf{Speech utterance embedding models:} to validate our method in the speech modality, we integrate it with three distinct speech embedding models.
First, we employ the SAMU-XLSR model \cite{khurana2022samu}, which leverages LaBSE as its teacher during the training. 
Next, we utilize SONAR, which supports both text and speech modalities, using its French speech encoder in this context.
Lastly, we make use of SENSE \cite{mdhaffar2025sensemodelsopensource}, which is trained using BGE-M3 as a teacher. \\
%Lastly, we develop a new model that follows a similar approach to SAMU-XLSR, but with two key modifications: BGE-M3 as the teacher model and w2v-BERT~2.0~\cite{seamlessm4t2023} for speech encoding. The CommonVoice 19 dataset is used to train this model~\cite{ardila2020common}.\\
%Our choice of these three specific models for both text and speech embeddings is motivated by the availability of matching speech versions. This ensures consistency in our evaluation across both modalities.\\
\textbf{Modified seq2seq model:} To achieve our primary objective of creating an abstractive 
summarization model for French, we chose BARThez \cite{eddine2021barthez}, a large-scale pretrained seq2seq model specifically tailored for French.
With its BART-based architecture and 
a low computational footprint of 165\,M parameters, BARThez has demonstrated strong performance in various generative tasks. 
However, we encountered an obstacle when attempting to integrate our sentence embedding models with BARThez: the expected input size for the BART model is 768 dimensions, whereas our embedded representations vary in size. To overcome this issue, we introduced a linear projection 
layer followed by a GeLU activation function to normalize the inputs. Notably, LaBSE embeddings are exempt from this transformation due to their pre-aligned dimensionality.
%\textbf{Adaptation Module} The BARThez model expects input sequences composed of vectors with a dimensionality of 768. However, the sentence embedding models we integrate may produce vectors of different sizes. To align these representations with BARThez's expected input space, we introduce a linear projection layer followed by a GeLU activation function, except for LaBSE embeddings, which already match BARThez's dimensionality and thus require only the activation function.

%\vspace{0.5em}

In the following, our model will be referred to as \textbf{SBARThez} for Semantic-BARThez. To further distinguish between variants that incorporate different sentence embedding models, we will append a suffix 
to the base model name (e.g., SBARThez-BGE).
%(Example: SBARThez-BGE when the BGE-M3 embeddings are used).  

\subsection{Evaluation Metrics}
\label{sub:metrics}
Since our summarization method relies on high-level semantic abstraction, assessing its quality is challenging: one must gauge the depth of understanding and insight it conveys, not merely its surface-level presentation.
To address this, we use various automatic evaluation metrics that range different facets of summarization quality. 
We report the standard ROUGE-L \cite{lin2004rouge} and BertScore \cite{zhang2019bertscore} metrics. 
We also consider three supplementary metrics to assess 
the abstractiveness of the automatically generated summaries: \\
%\begin{itemize}
\textbf{ROUGE-1 precision with source document (P-R1)} \cite{lin2004rouge} measures the proportion of unigrams (individual words) in the summary that also appear in the source document. \\
\textbf{Extractive Fragment Coverage (EFC)}  \cite{grusky2018newsroom} evaluates how much of the summary is directly derived from the source text. 
Extractive fragments are defined as the set of shared sequences of tokens in the source document and the summary. 
This set is denoted \( \mathcal{F}(D,S) \), where \( D \) is the document and \( S \) is the summary. EFC is calculated by the proportion of summary words that belong to these shared extractive fragments:
$\text{EFC}(D,S) = \frac{1}{|S|} \sum_{f \in \mathcal{F}(D,S)} |f|\enspace .$ \\
%\[
%\text{EFC}(A,S) = \frac{1}{|S|} \sum_{f \in \mathcal{F}(A,S)} |f|
%\]
%\( \text{EFC}(A,S) = \frac{1}{|S|} \sum_{f \in \mathcal{F}(A,S)} |f| \)
\textbf{Extractive Fragment Density (EFD)}  \cite{grusky2018newsroom} measures the average length of the extractive fragment that each word in the summary belongs to, taking into account both the inclusion of numerous original words and the impact of word arrangement on 
meaning preservation.
Like EFC, EFD is based on extractive fragments, but uses a square of the fragment length: \\
$\text{EFD}(D,S) = \frac{1}{|S|} \sum_{f \in \mathcal{F}(D,S)} |f|^2 \enspace .$ \\
%\[
%\text{EFD}(A,S) = \frac{1}{|S|} \sum_{f \in \mathcal{F}(A,S)} |f|^2
%\]
%\( \text{EFD}(A,S) = \frac{1}{|S|} \sum_{f \in \mathcal{F}(A,S)} |f|^2 \)
%\end{itemize}
%An example of computing these copy metrics is given in Appendix \ref{}.
\textbf{Named Entity Hallucination Risk (NEHR)}, introduced by \citet{akani2023reducing}, is calculated as the proportion of entities in the generated summary that are not present in the source document. This metric is used in our study in order to estimate the possible hallucination involving Named Entities. 

\vspace{-0.35cm}
\subsection{Tasks and Datasets}
\label{sub:tasks}
%\vspace{-0.2cm}
%Our approach involves a multi-stage training process to adapt the model to sentence embeddings as input.
Our proposed summarization approach involves a multi-stage training.
The first stage uses the French MLSUM dataset \cite{scialom2020mlsum}, one of the largest news summarization datasets available in French. 
This step is essential for helping the model adapt to sentence embeddings as input. 
The second stage focuses on task-specific training, where we tailor our model's performance for three summarization tasks: \\
\textbf{Monolingual Text Summarization} (Text FR→Text FR): to assess our model's performance on this task, we take the model initially trained on MLSUM and further fine-tune it on the OrangeSum corpus \cite{eddine2021barthez}. \\
\textbf{Cross-lingual Text Summarization} (Text X→Text FR): to validate the cross-lingual effectiveness of our approach, the MLSUM-trained model is fine-tuned separately on the French-WikiLingua \cite{ladhak2020wikilingua} and French-CrossSum \cite{bhattacharjee2023crosssum} datasets, and evaluated on the corresponding cross-lingual test-sets. \\
\textbf{Speech Summarization} (Speech FR→Text FR): to test our approach on speech, the MLSUM-trained model is further fine-tuned on DECODA \cite{bechet2012decoda}, which is a French human-human spoken conversation dataset.
Each conversation consists of three elements: the audio recording, the manual transcript and a summary. 
Building upon the work of \cite{akani2024unified}, we use the same version of DECODA to ensure comparability.

Table \ref{datasets-french} shows the distribution of text-summary pairs across training, development, and test sets of the studied datasets. 
The statistics for MLSUM, WikiLingua, and CrossSum correspond to their French subsets (French documents paired with their corresponding French summaries).
\begin{table}[H]
  \centering
  \resizebox{0.4\textwidth}{!}{%
  \begin{tabular}{crrr}
    \hline
    \textbf{Dataset} & \multicolumn{1}{c}{\textbf{Train}} & \multicolumn{1}{c}{\textbf{Dev}} & \multicolumn{1}{c}{\textbf{Test}} \\
    \hline
    {MLSUM}  & {392,876}       &{16,059}           &{15,828}                           \\
    {OrangeSum} & {21,401}       &{1,500}           &{1,500} \\
    {DECODA} & {1,390}       &{396}           &{200}  \\
    {WikiLingua} & {43,423} &{6,193} &{12,405} \\
    {CrossSum} & {8,648} &{1,083} &{1,083} \\
    \hline
  \end{tabular}
  }
  %\caption{\label{datasets-french}
  %  Datasets distributions by number of documents.
  %}
  \caption{\label{datasets-french}
    Number of documents per dataset
  }
\end{table}

\vspace{-0.8cm}
\subsection{Training Details}
\label{sub:training}
%from here
%We trained our model using the AdamW optimizer, with a batch size of 16 and different learning rates for the pretrained seq2seq model and the projection layer. 
%The pretrained seq2seq model parameters were optimized with a learning rate of $1\times10^{-5}$, while the linear projection layer was trained with a larger learning rate of $1\times10^{-3}$, allowing faster adaptation of the randomly initialized weights. 
%A weight decay of $1\times10^{-5}$ was applied uniformly to regularize both components. We used the standard cross-entropy loss to train the model on this sequence prediction task. \\
% to here
%To adaptively adjust the learning rates during training, we employed the same scheduler for both optimizers, which reduces the learning rate by a factor of 0.5 if the validation loss does not improve for two consecutive epochs. 
%This scheduler helps avoid overfitting and stabilizes training, especially for the low-learning-rate pretrained backbone.

For each task, we follow the same two-stage learning strategy.
In the first training stage, the model is optimized using the AdamW optimizer, with a batch size of 16 and different learning rates for the pretrained seq2seq model and the projection layer. 
The pretrained seq2seq model parameters were optimized with a learning rate of $1\times10^{-5}$, while the linear projection layer was trained with a larger learning rate of $1\times10^{-3}$, allowing faster adaptation of the randomly initialized weights.
A weight decay of $1\times10^{-5}$ was applied uniformly to regularize both components. We used the standard cross-entropy loss to train the model on this sequence prediction task. 
In the second stage, we further fine-tune the model obtained at the end of the first stage.
%Further details on the training hyperparameters are provided in Appendix~\ref{sec:appendixA}.
%Further details are presented in the next section. 

%\section{Results and Discussion}
%\subsection{Monolingual Text Summarization}

%In what follows, we present the results of our experiments: Section \ref{sec:mono_text} covers monolingual text summarization (Text FR → Text FR), Section \ref{sec:cross_text} addresses cross-lingual text summarization (Text X → Text FR), and Section \ref{sec:speech_summ} focuses on speech summarization (Speech FR → Text FR).

We present the results of our experiments in the following sections: section \ref{sec:mono_text} covers monolingual text summarization, section \ref{sec:cross_text} addresses cross-lingual text summarization, and section \ref{sec:speech_summ} focuses on speech summarization.

%\vspace{-0.79cm}
\section{Monolingual Text Summarization}
\label{sec:mono_text}
%\subsection{Analysis of Entity Hallucinations}
\subsection{Preliminary Results}
We first evaluate the originally fine-tuned BARThez model on the OrangeSum test-set. 
This model serves as our baseline, representing the standard encoder-decoder architecture relying on tokenized input sequences. 
As shown in Table \ref{results-orange1}, the SBARThez-BGE model achieves the best performance among all SBARThez variants.
Despite its overall improved performance and results that are close to those of the baseline BARThez model in terms of Rouge-L and BertScore, the NEHR remains high for all SBARThez variants. For example, SBARThez-BGE reaches a NEHR of 58.52\%, indicating that more than half of its named entities are hallucinated.
This limitation is likely due to the highly semantic and abstractive nature of the SBARThez approach, which does not explicitly preserve lexical forms.
To address this issue, we suggest including a Named Entity Injection (NEI) mechanism in the SBARThez architecture, as detailed in the following section.

\begin{table}[H]
  \centering
  \resizebox{0.4\textwidth}{!}{%
  \begin{tabular}{cccc}
    \hline
    \textbf{Model} &\textbf{R-L~$\uparrow$} &\textbf{BertS~$\uparrow$} &\textbf{NEHR~$\downarrow$} \\
    \hline
    {Ground Truth} & 100 & 100 &{34.01} \\
    \hline
    {BARThez} & \textbf{22.57} & \textbf{27.95} &\textbf{26.88}\\
    \hline
    {SBARThez-BGE} & \underline{19.24} & \underline{24.61} &\underline{58.52}\\
    {SBARThez-LaBSE} & 17.53 & 22.83 &{71.79}\\
    {SBARThez-SONAR} & 19.10 & 24.58 &{59.23}\\
    \hline
  \end{tabular}
  }
  %\caption{\label{results-orangesum-nehr}
  %  Results on the OrangeSum test-set.
  %}
  \caption{\label{results-orange1}
    Results on the OrangeSum test-set. Best results are \textbf{bold}, and second-best results are \underline{underlined}. BertS refers to BertScore.
}
\end{table}

\subsection{Named Entity Injection Mechanism}
\begin{figure}[ht!]
  \centering
  \includegraphics[width=0.9\columnwidth, keepaspectratio]{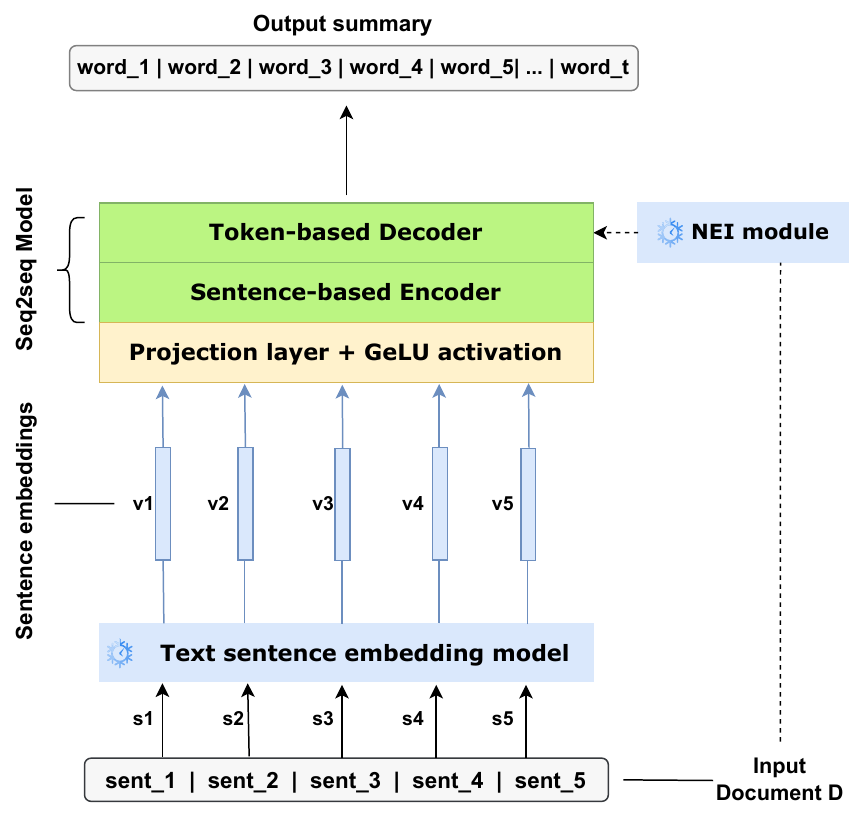}
  \caption{The sentence-based summarization model architecture with the NEI mechanism}
  \label{fig:model_archi}
\end{figure}

To address the issue of hallucinated named entities, we incorporate a NEI mechanism into our approach, as illustrated in Figure \ref{fig:model_archi}. 
Named entities are extracted from the input document using a Named Entity Recognition (NER) system, then tokenized using the seq2seq model's tokenizer %(\texttt{NE\_tokens})
and appended to the decoder inputs.
During training in both stages, the named entity tokens are incorporated into the decoder block of the BARThez seq2seq model, alongside the standard decoder's input tokens and the encoder’s hidden states. 
At inference time, the decoder is initialized with these tokens as start tokens, allowing the generative process to utilize them in conjunction with the sentence-level embeddings of the input document.
This allows the summarization model to leverage both the semantic sentence embeddings on the encoder side and the explicit named entities on the decoder side. 
The NER model employed in our study is camembert-ner\footnote{\url{https://huggingface.co/Jean-Baptiste/camembert-ner}}, a French NER model fine-tuned from CamemBERT \cite{martin2020camembert} on the wikiner-fr dataset \cite{nothman2013learning}.
The NER system categorizes extracted entities into four classes: PER (persons), ORG (organizations), LOC (locations), and MISC (miscellaneous). 
Following the extraction, we apply a preprocessing step where entities with a confidence score below 0.9 are filtered out.

%\begin{figure*}[ht!]
%  \centering
%  \includegraphics[width=0.79\textwidth, keepaspectratio]{LREC2026 Author's kit/figures/nouv_schema_sbarthez.drawio3.pdf}
  %\includegraphics[width=0.85\textwidth, keepaspectratio]{LREC2026 Author's kit/figures/nouv_schema_sbarthez.drawio3.pdf}
%  \caption{Architecture of our proposed abstractive summarization model with Named Entity Injection}
%  \label{fig:model_archi}
%\end{figure*} 

\vspace{-0.5cm}
\subsection{Results using the NEI Mechanism}
To assess the effectiveness of the NEI module, we conduct a series of experiments to identify the best named entity (NE) prompt for injection.
Among the prompts tested, we identified one structure, which achieved the highest BertScore on OrangeSum: ``$(E_1, T_1), (E_2, T_2), \ldots, (E_n, T_n).$''
This prompt lists each named entity \(E_i\) alongside its type \(T_i\), where \( T_i \in \{\text{PER}, \text{ORG}, \text{LOC}, \text{MISC}\} \), with duplicates removed.It is then tokenized using the BARThez tokenizer to produce NE tokens for the decoder input.

\begin{table}[H]
  \centering
  \resizebox{0.4\textwidth}{!}{%
  \begin{tabular}{cccc}
    \hline
    \textbf{Model} &\textbf{R-L~$\uparrow$} &\textbf{BertS~$\uparrow$} &\textbf{NEHR~$\downarrow$} \\
    \hline
    {Ground Truth} & 100 & 100 &{34.01} \\
    \hline
    {BARThez} & \textbf{22.57} & \textbf{27.95} &\textbf{26.88}\\
    \hline
    {SBARThez-BGE} & \underline{20.22} & \underline{26.05} & \underline{34.16}\\
    {SBARThez-LaBSE} & 18.56 & 24.24 &{38.41}\\
    {SBARThez-SONAR} & 19.40 & 25.37 &{36.31}\\
    \hline
  \end{tabular}
  }
  %\caption{\label{results-orangesum-nehr}
  %  Results on the OrangeSum test-set.
  %}
  \caption{\label{results-orangesum-nehr}
    Results on the OrangeSum test-set, with SBARThez models using the NEI module.
}
\end{table}

As shown in Table~\ref{results-orangesum-nehr}, NE injection improves the performance of all variants of our approach compared to the results presented in Table \ref{results-orange1}. The best-performing model remains the token-based BARThez model, followed by SBARThez-BGE with NE injection, which achieves the highest scores among all SBARThez variants. 

Regarding named entity hallucination, the ground truth summaries exhibit a NEHR of 34.01\%, indicating that 34\% of the named entities in the summaries do not appear in the original documents.
This suggests that human paraphrasing during summary creation may be a contributing factor, such as using ``France'' to refer to ``French Republic''.
It also highlights a limitation of NEHR, since it relies on exact matching and is therefore overly strict.
Nevertheless, it provides a useful proxy for estimating the risk of named entity hallucination.

BARThez achieves the lowest NEHR, indicating stronger faithfulness to the source text in terms of NE generation.
However, this approach comes at the cost of limiting the model's ability to generalize or rephrase entities more naturally and expressively. While this may enhance factual accuracy, it may also restrict the model's creative potential.
In contrast, our SBARThez models enhanced with NE injection show a significant and consistent reduction in  named entity hallucination.
Specifically, SBARThez-BGE originally hallucinated over half of the named entities (Tab \ref{results-orange1}), but with the addition of the NE injection module, the NEHR decreased to 34\%, comparable to the ground truth.

%For all the upcoming experiments using NE injection, we will be using the P4 prompt. 
% SUBSECTION ABOUT ABSTRACTIVENESS ?
%\vspace{-0.5cm}

%\subsubsection{Analysis of Abstractiveness}
\subsection{Analysis of Abstractiveness}

Table \ref{results-all} presents the results of the SBARThez variants with NE injection. All token-based encoding models are fine-tuned on OrangeSum. The evaluation includes Rouge-L and BertScore, as well as the copy rate metrics to assess the abstractiveness of the generated summaries.

The SBARThez-BGE model outperforms other variants in terms of Rouge-L and BertScore, while also surpassing mT5-small. Its performance is remarkably close to that of mBART-50 and BARThez, which lead on these metrics.
One notable aspect of the token-level encoding models is their extremely high P-R1, EFC, and EFD values, indicating an over-reliance on copying from the source document.
For example, the P-R1 value shows that BARThez uses over 90\% of the words in its generated summaries that are present in the source document.
In contrast, SBARThez models exhibit lower copying rates and demonstrate a remarkable ability to match human summaries in abstractiveness while maintaining competitive performance on R-L and BertScore metrics.
Besides, SBARThez summaries tend to be shorter than ground-truth summaries, according to the calculated length. This suggests the model's ability to condense complex information into concise summaries.

%\textbf{Results with Best NE-prompts on different systems}
%\begin{table*}
%  \centering
%  \begin{tabular}{cccccccc}
%    \hline
%    Model &R-L &BertScore &Fact-CC &P-R1 &EFC &EFD &Fact-CC\\
%    \hline
%    {Ground Truth}  &{100} &{100/100} &{x}  &{79.05} &{76.39} &{2.65} &{x}                         \\
%    \hline
%    {BARThez}  &{22.57} &{27.95/73.00} &{x} &{92.22} &{90.93} &{7.17}    &{x}                      \\
%    \hline
%    {SBARThez-BGE w/ NE}  &{19.84} &{25.74/72.17} &{x} & {79.30}  &{76.26} &{1.81} &{x}                            \\
%    {SBARThez-LaBSE w/ NE} &{18.56} &{24.24/71.60} &{x} &{75.31} &{72.81} &{1.64}      &{x}                     \\
%    {SBARThez-SONAR w/ NE} &{19.40} &{25.37/72.03} &{x} &{79.32} &{76.41} &{1.79}      &{x}                     \\
%    \hline
%  \end{tabular}
%  \caption{\label{results-all}
%    Results with Best NE-prompts on different systems on OrangeSum
%  }
%\end{table*}

\begin{table}[h]
\centering
%\setlength{\tabcolsep}{2pt}  % default is 6pt
%\scriptsize % Smaller font
%\setlength{\tabcolsep}{2pt} % Reduce horizontal space
%\renewcommand{\arraystretch}{1.2} % Reduce row height
%\resizebox{\columnwidth}{!}{%
\resizebox{0.5\textwidth}{!}{%
\begin{tabular}{l cc ccc c}
\toprule
\multirow{2}{*}{Model}
& \multicolumn{2}{c}{\textbf{Ground Truth}} 
& \multicolumn{3}{c}{\textbf{Source Document}} 
& \multirow{2}{*}{Length} \\
\cmidrule(lr){2-3} \cmidrule(lr){4-6} & R-L~$\uparrow$ & BertS~$\uparrow$ & P-R1~$\downarrow$ & EFC~$\downarrow$ & EFD~$\downarrow$ & \\
\midrule
\multicolumn{7}{l}{\textbf{Token-Encoding}} \\
%\multicolumn{8}{l}{\textbf{Token-Level Encoding}} \\
%\multicolumn{1}{l}{\textbf{ \space \space \space \space \space \space \space \space \space \space \space \space \space \space \space Token-Level based}} & \multicolumn{2}{l}{\textbf{Encoding}} \\
BARThez & \underline{22.57} & \textbf{27.95} & 91.65 & 89.11 & 6.66 & 24.71 \\
flanT5-base & \textbf{22.61} & \underline{27.54} & 94.00 & 92.37 & 8.60 & 27.87 \\
mT5-small & 20.22 & 24.06 & 95.22 & 95.39 & 9.99 & 25.88 \\
mBART-50 & 22.05 & 26.76 & 93.13 & 90.96 & 8.63 & 27.72 \\
\hline
%\multicolumn{8}{l}{\textbf{Sentence-Level Encoding}} \\
%\multicolumn{1}{l}{\textbf{\space \space \space \space \space \space \space \space \space \space \space \space Sentence-Level based}} & \multicolumn{2}{l}{\textbf{Encoding (ours)}} \\
%\multicolumn{1}{l}{\textbf{Sentence-Level Encoding}} & \multicolumn{2}{l}{\textbf{ (ours)}} \\
\multicolumn{7}{l}{\textbf{Sentence-Encoding}} \\
SBARThez-BGE & 20.22 & 26.05 & 78.54 & \underline{72.30} & 1.67 & 20.67 \\
SBARThez-LaBSE & 18.56 & 24.24 & \textbf{74.64} & \textbf{68.71} & \textbf{1.45} & 21.90 \\
SBARThez-SONAR & 19.40 & 25.37 & \underline{78.52} & 72.70 & \underline{1.60} & 20.93 \\
\hline
Ground Truth & 100.0 & 100.0 & 78.36 & 72.84 & 2.41 & 32.13 \\
\hline
\end{tabular}%
}
\label{results-all}
\caption{
Comparison of summarization methods on OrangeSum in terms of abstractiveness, accuracy, and summary length (average words excluding punctuation).
}
\end{table}

%\subsection{Cross-lingual Summarization}
%\vspace{-1.1cm}
\section{Cross-lingual Text Summarization}
\label{sec:cross_text}
\vspace{-0.1cm}
\subsection{Evaluating Model Robustness across Multiple Languages}
In this section, we evaluate the cross-lingual abstractive summarization capabilities of our approach. 
The goal is to generate French summaries from source documents written in various languages.
To this end, we fine-tuned the MLSUM-trained model on the French WikiLingua dataset, which contains source documents and summaries in French.
This finetuning improves the model's relevance to the dataset's specific characteristics.
%This fine-tuning step enhances the model’s ability to adapt to the specific characteristics and requirements of the dataset.
The NE injection module is not used, as WikiLingua consists of how-to guides, which contain very few named entities.
Indeed, our analysis of the training set source documents revealed an average of only 1.37 named entities per document.
%For evaluation, we use the cross-lingual test-set with various X→Y language pairs, where X represents the language of the source document and Y is the language of the summary.
%For evaluation, We filter the test-set to include only X→FR pairs, where X is one of the following languages: English, Spanish, Portuguese, Italian, German, or Dutch. 
%We then align the extracted test subsets based on their French summaries to construct a single unified test-set across these languages. 
%This allows for a consistent comparison of cross-lingual summarization results.

For evaluation, we used a unified WikiLingua test-set constructed as the intersection of test sets of the studied source languages (French, English, Spanish, Italian, Portuguese, Dutch, German), aligned through their corresponding French summaries, resulting in 2,166 common test samples.

We observe from table \ref{results-wikilingua} that performance across languages for each SBARThez model remains relatively consistent.
For example, the BertScore is around 28 for EN, ES, IT, PT and NL with SBARThez-BGE, indicating that summarization quality does not significantly vary between these languages.
The same trend holds for the other two SBARThez variants. 
However, when comparing these results to the FR→FR setup, we notice a drop in performance.
% particularly for SBARThez-BGE, which loses approximately 4 points in both BertScore and R-L. 
%In contrast, SBARThez-SONAR and SBARThez-LaBSE only show a loss of about 2 points.
This may be because SBARThez models were fine-tuned on the French WikiLingua dataset, making them more suitable for processing text embeddings derived from French data.
%However, SBARThez-SONAR shows the best results in the cross-lingual evaluation, reaching a BertScore of 30 in most languages, followed by SBARThez-BGE, while SBARThez-LaBSE yields the lowest scores.

%\setlength{\tabcolsep}{4pt}
\begin{table}[h]
  \centering
  \resizebox{0.47\textwidth}{!}{%
  \begin{tabular}{lcccccc}
    \toprule
    \multirow{2}{*}{} 
    & \multicolumn{2}{c}{\makecell{SBARThez\\BGE}} 
    & \multicolumn{2}{c}{\makecell{SBARThez\\LaBSE}}
    & \multicolumn{2}{c}{\makecell{SBARThez\\SONAR}} \\
    \cmidrule(lr){2-3} \cmidrule(lr){4-5} \cmidrule(lr){6-7}
    & \textbf{R-L~$\uparrow$} & \textbf{BertS~$\uparrow$} & \textbf{R-L~$\uparrow$} & \textbf{BertS~$\uparrow$} & \textbf{R-L~$\uparrow$} & \textbf{BertS~$\uparrow$} \\
    \midrule
    FR & {24.18} & {32.80} & 20.17 & 26.32 & {23.18} & {32.21} \\
    \midrule
    EN & {20.29} & {28.30} & 18.87 & 24.89 & {21.56} & {30.03} \\
    ES & {20.06} & {28.70} & 18.80 & 24.25 & {21.54} & {29.62} \\
    IT & {19.89} & {28.63} & 18.25 & 24.38 & {20.82} & {29.47} \\
    PT & {19.70} & {28.50} & 18.56 & 25.22 & {21.19} & {30.07} \\
    NL & {20.23} & {28.35} & 19.16 & 25.38 & {21.94} & {30.10} \\
    DE & {19.52} & {27.76} & 18.71 & 25.16 & {21.71} & {30.47} \\
    \bottomrule
  \end{tabular}
  }
  \caption{\label{results-wikilingua}
    Results on WikiLingua commun test-set.
  }
\end{table}
%\vspace{-\baselineskip}
%\vspace{-0.5cm}

%\begin{table}[H]
%  \centering
%  \begin{tabular}{cccccc}
%    \hline
%    Model &LNG &R-L &BertScore \\
%    \hline
%    \multirow{6}{*}{SBARThez-BGE} &{FR}      &{24.18} &{32.80}                         \\
%    &{EN}      &{20.29} &{28.30}                         \\
%    &{ES}     &{20.06}  &{28.70}                     \\
%    &{IT}     &{19.89}  &{28.63}              \\
%    &{PT}     &{19.70}  &{28.50}   \\
%    &{NL}     &{20.23}  &{28.35}   \\
%    &{DE}     &{19.52}  &{27.76}   \\
%    \hline
%    \multirow{6}{*}{SBARThez-LaBSE}&{FR}      &{20.17} &{26.32} \\
%    &{EN}      &{18.87} &{24.89}                         \\
%    &{ES}     &{18.80}  &{24.25}                     \\
%    &{IT}     &{18.25}  &{24.38}              \\
%    &{PT}     &{18.56}  &{25.22}   \\
%    &{NL}     &{19.16}  &{25.38}   \\
%    &{DE}     &{18.71}  &{25.16}   \\
%    \hline
%    \multirow{6}{*}{SBARThez-SONAR}&{FR}      &{23.18} &{32.21} \\
%    &{EN}      &{21.56} &{30.03}                         \\
%    &{ES}     &{21.54}  &{29.62}                     \\
%    &{IT}     &{20.82}  &{29.47}              \\
%    &{PT}     &{21.19}  &{30.07}   \\
%    &{NL}     &{21.94}  &{30.10}   \\
%    &{DE}     &{21.71}  &{30.47}   \\
%    \hline
%  \end{tabular}
%  \caption{\label{results-wikilingua}
%    Results on the WikiLingua commun test-set
%  }
%\end{table}

%\vspace{-0.3cm}
\subsection{Performance in High and Low-Resource Language Settings}

As in the previous section, the goal here is to generate French summaries from source documents written in both high- and low-resource languages.
We use the MLSUM-trained model, retraining it solely on the French portion of CrossSum (French documents paired with French summaries) to improve its relevance to the dataset's specific requirements.
Due to the scarcity of accurate NER systems 
for low-resource languages, we opt out of using the NE injection module.
We employ SBARThez-BGE and compare its performance against two baselines: \\
\textbf{Translate-Then-Summarize (TTS)} involves translating source documents into French using the M2M-100 1.2B Machine Translation (MT) model \cite{fan2021beyond}, followed by summary generation using BARThez fine-tuned on French CrossSUM. \\
\textbf{A multilingual large language model (LLM)}, specifically 
LLaMA-8B, is prompted in a zero-shot setting to produce French summaries.

The results are presented in Tables~\ref{results-crosssum-high} and~\ref{results-crosssum-low}.
The evaluation set, denoted as Test-Dev, combines the test and development sets.
The number of samples is displayed alongside each result.
Our approach excels in high-resource languages according to BertScore, outperforming other methods in English, Ukrainian and Japanese while closely matching the TTS method's performance in other high-resource languages.
Notably, our SBARThez-based method outperforms both TTS and LLM approaches across all low-resource languages, with exceptions for Urdu (UR) and Sinhala (SI).

%A key point to emphasize is that languages such as Igbo (IG), Kirundi (RN), and Pidgin (PCM) are not supported by the BGE embedding model. 
A key point to emphasize is that languages such as Igbo (IG), Kirundi (RN), and Pidgin (PCM) were not seen in the training of the BGE embedding model. 
Despite this limitation, our SBARThez-BGE approach still achieves strong results and outperforms all other methods on these languages.
In contrast, for Kirundi and Pidgin, we were unable to find suitable MT systems to translate these languages into French, rendering the TTS approach ineffective in the absence of available MT models. 
%This highlights a key limitation of TTS methods: their heavy dependence on the availability and quality of MT systems.
%For instance, even though a MT system exists for Gujarati (GU), the TTS approach only reaches a very low BertScore of 03.09. 

Overall, these findings demonstrate the significant advantage of our approach for cross-lingual summarization, particularly in low-resource languages, where it maintains strong performance even in the absence of NER resources or embedding model support for the source language.

\begin{table*}[h]
  \centering
  \resizebox{0.85\textwidth}{!}{%
  \begin{tabular}{lcccccccccc}
  %\cmidrule(lr){2-11}
  \hline
   & EN & AR & RU & UK & KO & JA & VI & ID & SW & HI \\
  \hline
  %SBARThez-BGE & 19.96/29.40 & 20.21/29.83 & 19.87/29.15 & 18.77/29.91 & 18.33/28.59 & 12.63/20.01 & 20.44/28.92 & 20.09/30.64 & 20.59/30.95 & 17.92/25.61 \\
  SBARThez-BGE (ours) & \textbf{29.40} & \underline{29.83} & \underline{29.15} & \textbf{29.91} & \underline{28.59} & \textbf{33.25} & \underline{28.92} & \underline{30.64} & \underline{30.95} & \underline{25.61} \\
  MT+BARThez  & \underline{29.11} & \textbf{30.71} & \textbf{30.05} & \underline{28.89} & \textbf{28.68} & 15.65 & \textbf{30.18} & \textbf{31.62} & \textbf{31.33} & \textbf{26.49} \\
  LLaMA-8B  & 25.00 & 26.33 & 24.61 & 25.61 & 24.19 & \underline{29.23} & 26.12 & 25.61 & 21.48 & 20.68 \\
  \hline
  \#Test-Dev Samples & 377 & 197 & 160 & 143 & 52 & 54 & 88 & 204 & 174 & 119 \\
  \hline
  \end{tabular}
  }
  \caption{\label{results-crosssum-high}
    Results on CrossSum using BertScore~$\uparrow$ – \textbf{High-Resource Languages}
  }
\end{table*}

\begin{table*}[h]
  \centering
  \resizebox{\textwidth}{!}{%
  \begin{tabular}{lcccccccccccccccc}
  \hline
   & PA & UR & PS & MR & YO & SO & AZ & TA & MY & SI & GU & IG & RN & PCM \\
  \hline
  SBARThez-BGE (ours) & \textbf{18.40} & 17.19 & \textbf{28.38} & \textbf{25.81} & \textbf{30.75} & \textbf{28.52} & \textbf{29.44} & \textbf{25.72} & \textbf{26.03} & \underline{25.98} & \textbf{25.11} & \textbf{26.77} & \textbf{26.14} & \textbf{29.81} \\
  MT+BARThez  & \underline{16.05} & \underline{17.67} & \underline{27.42} & \underline{24.37} & 16.25 & 05.28 & 22.54 & 15.24 & 06.44 & \textbf{27.09} & 03.09 & 05.45 & - & - \\
  LLaMA-8B   & 14.62 & \textbf{21.61} & 25.99 & 19.89 & \underline{23.96} & \underline{21.64} & \underline{24.22} & \underline{15.32} & \underline{21.65} & 24.03 & \underline{19.38} & \underline{17.92} & \underline{22.57} & \underline{25.53} \\
  \hline
  \#Test-Dev Samples & 45 & 152 & 36 & 46 & 33 & 100 & 42 & 84 & 17 & 19 & 46 & 20 & 144 & 88 \\
  \hline
  \end{tabular}
  }
  \caption{\label{results-crosssum-low}
    Results on CrossSum using BertScore~$\uparrow$ – \textbf{Low-Resource Languages}.
  }
\end{table*}

%\vspace{-0.2cm}
\section{Speech Summarization}
\label{sec:speech_summ}
%\vspace{-0.23cm}
\subsection{Experimental Setup and Results}
\label{sec:speech_results}
Three experiments were conducted to adapt the MLSUM-trained model to the speech modality.
First, the model was fine-tuned on manual transcriptions from DECODA, resulting in a text-based model.
Second, we fine-tuned the model on DECODA speech data, using three separated speech utterance embeddings generated by either SONAR, SAMU-XLSR, or SENSE.
Finally, we experimented with fine-tuning the model using both speech and text modalities by merging and shuffling their text and speech embeddings. \\
\textbf{Baselines:} For comparative analysis, we also fine-tuned BARThez on the manual transcriptions of DECODA and integrated it with two different Automatic Speech Recognition (ASR) systems. \textbf{Cascaded\_1} uses Whisper Tiny for ASR and BARThez for summarization, resulting in a Word Error Rate (WER) of 64.13\% on the test set. In contrast, \textbf{Cascaded\_2} leverages Whisper Large for ASR and BARThez for summarization, achieving a WER of 26.01\% on the test set. \\
\textbf{NE extraction:} To enable the NE injection module, we fine-tuned the Whisper-small model \cite{radford2023robust} on DECODA to minimize WER.
%In order to use the NE injection module, we needed an ASR system.
%To achieve this, we fine-tuned the Whisper-small model \cite{radford2023robust} on DECODA, with the goal of minimizing the WER.
The resulting ASR model was integrated with the SpeechBrain toolkit~\cite{ravanelli2024open} to transcribe speech and then extract named entities from the transcriptions using the NER model. \\
\textbf{Audio segmentation:} To segment the audio files and extract the sequence of speech utterance embeddings used as input to the SBARThez model, we rely on the ground-truth time boundaries provided in DECODA.
Each audio file in the corpus is divided into segments with annotated start and end times.
Accordingly, if an audio file contains 8 segments, we generate 8 corresponding speech embeddings; one for each segment. \\
\textbf{Results Analysis: } Results presented in Table~\ref{results-decoda1} reveal that all 
SBARThez variants exhibit their best performance on the speech test-set when trained with both text and speech modalities.
This indicates that training with speech embeddings enhances the model’s performance on speech.
%This trend holds true regardless of whether the evaluation input is text or speech.
In addition, SBARThez-BGE achieves a BertScore of 35.07 on the text test-set and 35.05 on the speech test-set, demonstrating consistent performance across modalities.
%Although the models are not fine-tuned on speech, they still perform comparably on speech inputs.
%However, models trained with speech utterance embeddings achieve superior summarization results compared to models trained solely on text and applied to speech inputs during inference.
As shown in the same table, WER has a clear impact on the performance of cascaded approaches. 
For example, when switching from Whisper Large (lower WER) to Whisper Tiny, the BertScore drops significantly, from 38.54 to 32.55.
In comparison, while SBARThez-BGE and SBARThez-SONAR outperform the Cascaded\_1, they are still behind the Cascaded\_2 system. 
This performance gap may be explained by the use of a generic NER system that can only extract PER, ORG, and LOC entities, while DECODA mainly contains domain-specific entities such as phone numbers, prices, and transport types, which are not captured by the generic NER system. 
\vspace{-0.3cm}
%Future work will focus on collecting data to train a specialized NER system tailored to DECODA and integrating it into our approach for summary generation.

%Overall, our model represents a near end-to-end approach, generating summaries directly from speech utterance embeddings without intermediate transcriptions. 
%This opens promising avenues for future research in end-to-end speech summarization.

\begin{table}[h!]
%\begin{table}[H]
  \centering
  \resizebox{0.42\textwidth}{!}{%
  \begin{tabular}{ccccc}
    \hline
    \textbf{Model} & \textbf{\makecell{Train\\Data}} & \textbf{\makecell{Test\\Data}} & \textbf{R-L~$\uparrow$} & \textbf{\makecell{BertS~$\uparrow$}} \\
    \hline
    BARThez & Text & Text & 29.11 & 38.90 \\
    Cascaded\_1 & - & Speech & 23.43 & 32.55 \\
    Cascaded\_2 & - & Speech & \textbf{27.82} & \textbf{38.54} \\
    \hline
    \multirow{4}{*}{\makecell{SBARThez\\BGE}} & Text & Text   & 24.64 & 35.07 \\
                              & Text & Speech & 22.86 & 32.05 \\
                              & Speech & Speech & 23.99 & 34.64 \\
                              & Both & Speech & \underline{24.43} & \underline{35.05} \\
    \hline
    \multirow{4}{*}{\makecell{SBARThez\\LaBSE}} & Text & Text   & 22.26 & 32.39 \\
                              & Text & Speech & 21.03 & 30.28 \\
                              & Speech & Speech & 21.09 & 31.03 \\
                              & Both & Speech & 22.05 & 31.39 \\
    \hline
    \multirow{4}{*}{\makecell{SBARThez\\SONAR}} & Text & Text   & 24.52 & 34.89 \\
                              & Text & Speech & 20.88 & 32.44 \\
                              & Speech & Speech & 23.58 & 32.73 \\
                              & Both & Speech & 23.34 & 33.50 \\
    \hline
  \end{tabular}
  }
  \caption{\label{results-decoda1} Results on the DECODA test-set.}
\end{table}

%\vspace{-0.9cm}
%\subsection{Impact of audio segmentation on performance}
\subsection{Impact of Audio Segmentation}

In real-world scenarios, the ground-truth (GT) audio segmentations used to obtain the results in subsection \ref{sec:speech_results} are not available.
Therefore, we use automatic audio segmentation methods and analyze their impact on the effectiveness of speech summarization. These methods include:

\textbf{Content-aware Segmentation}:  The boundaries of an audio segment are determined by the acoustic or linguistic properties of the signal. We evaluate two methods: \textbf{(1) Voice Activity Detection (VAD)} detects the presence or absence of human speech, segmenting the audio where speech begins and ends. We use the Silero-VAD tool \cite{SileroVAD}. \textbf{(2) Speaker Diarization (Spk DIAR)} divides audio into homogeneous segments based on speaker identity, thereby segmenting the audio at speaker turns. We use the pyannote toolkit \cite{bredin2023pyannote}.

\textbf{Time-driven Segmentation}: The audio is divided into fixed-length segments, regardless of speech or speaker content. This simple approach relies solely on duration. We evaluate three variants with segment lengths of 3, 5, and 8 seconds.

We evaluate the fine-tuned SBARThez-BGE model, trained on manual transcriptions of DECODA, using different segmentation strategies (Table \ref{tab:seg-results}).
The results show that all segmentation methods maintain summary quality, with Rouge-L and BertScore values comparable to those obtained using GT segmentation.
Among automatic methods, segmentation into 5-second segments achieves the best performance after the GT segmentation, while 8-second segments also yield results close to the GT baseline.
This indicates that our approach remains effective even when using longer audio segments.
Overall, our method demonstrates consistent performance across various segmentation approaches, including simple time-based ones, making it valuable for real-world applications.

\begin{table}[h]
\centering
\resizebox{0.4\textwidth}{!}{%
\begin{tabular}{llcc}
\toprule
\textbf{Category} & \textbf{Method} & \textbf{R-L~$\uparrow$} & \textbf{BertS~$\uparrow$} \\
\midrule
\textsl{GT Segmentation} & \textsl{-} & \textsl{24.64} & \textsl{35.07} \\
\midrule
\multirow{2}{*}{\makecell{Content-aware\\Segmentation}}
& VAD & 23.51 & 34.43 \\
& Spk DIAR &\underline{24.18} & \underline{34.95} \\
%& Hybrid & 23.70 & 33.84 \\
\midrule
\multirow{3}{*}{\makecell{Time-driven\\Segmentation}}
& 3sec fixed & 23.83 & 34.23 \\
& 5sec fixed & \textbf{24.42} & \textbf{34.98} \\
& 8sec fixed & 23.80 & 34.22 \\
\bottomrule
\end{tabular}
}
\caption{Segmentation results with SBARThez-BGE. Evaluation is made on speech segments using different segmentation methods.}
\label{tab:seg-results}
\end{table}

\vspace{-0.8cm}
\section{Conclusion}
%\vspace{-0.3cm}

In this paper, we introduce SBARThez, a novel abstractive summarization approach that leverages semantic sentence-level embeddings, moving beyond traditional token-level encoding.
We address the issue of hallucinated named entities with a Named Entity Injection mechanism.
Our experimental results show that SBARThez produces more abstractive summaries than token-based models, while maintaining competitive performance and surpassing strong baselines in the cross-lingual setting, particularly for low-resource languages.
Our approach is also highly versatile: it can be applied to both text and speech inputs, and supports multilingual source documents.
This allows for efficient generation of high-quality summaries from multiple languages without requiring additional training or adaptation.
This work opens up promising avenues for future research in end-to-end speech summarization and cross-lingual summarization.

%\vspace{-0.5cm}

\section{Limitations}
% ---- IDEAS 
%- We could not test the NE injection module in the cross-lingual section \\
%- We would like to train a NER model for DECODA to evaluate on speech and improve performance\\
%- This work uses the BARThez seq2seq model for summary generation, but it would be possible to use other seq2seq models and evaluate the performance based on it \\
%- We focus on generating french summaries in this work, but it can be extensible to generate english (or any other languages) summaries\\
%- We would be interested on evaluating our approach on low-resource languages\\ 

%Our experiments are based on the BARThez architecture.
%Future work could explore the impact of alternative architectures, such as T5, mBART, or GPT-style models, to determine whether using a larger or more sophisticated pretrained model would improve performance. 

Our study focuses on generating French abstractive summaries. 
While the methodology could extend to other languages, including English or low-resource ones, further validation is needed to confirm its generalizability.
Furthermore, although we demonstrate the effectiveness of our approach for summarization, its applicability to other NLP tasks, such as question answering, paraphrasing, or text simplification, remains to be explored. 
Generalizing the method across different tasks could provide broader insights into its versatility and robustness. 
Finally, one key limitation in this study is the lack of suitable factual consistency metrics for French.
Existing open-source metrics are primarily designed for English, and while G-Eval \cite{liu2023g} is currently one of the best-performing metrics with GPT-based models, its reliance on closed-source, high-cost models prevented us from using it in our evaluation.
Despite these limitations, we believe this work introduces a promising new approach to abstractive summarization and makes a meaningful contribution to ongoing research in the field. 

\section{Acknowledgments}
This work used HPC resources from GENCI-IDRIS: grants AD011015509R1, A0181012551, and AD011016301.

\nocite{*}
\section{Bibliographical References}\label{sec:reference}

\bibliographystyle{lrec2026-natbib}
\bibliography{lrec2026-example}

\bibliographystylelanguageresource{lrec2026-natbib}
%\bibliographylanguageresource{languageresource}

\end{document}